\documentclass[10pt, a4paper]{article}

\usepackage{times}
\usepackage{latexsym}
\usepackage{amsmath,amsfonts,amssymb,amsthm}
\usepackage{comment}
\usepackage{booktabs}
\usepackage{caption}
\usepackage{subcaption}
\usepackage[ruled,noline,noend,lined]{algorithm2e}
\usepackage{multirow}
\usepackage{array}
\usepackage{xcolor}
\usepackage{soul}
\usepackage{float}
\usepackage{graphicx}
\usepackage{enumitem}
\usepackage{microtype}
\usepackage{tipa}
\usepackage{times}
\usepackage{latexsym}
\usepackage[T1]{fontenc}
\usepackage[utf8]{inputenc}
\usepackage{microtype}
\usepackage{inconsolata}

\usepackage[]{lrec-coling2024} 

\title{SpeechAlign: a Framework for Speech Translation Alignment Evaluation}

\name{Belen Alastruey$^{1*}$\thanks{\hspace{1.3mm} Equal contribution.} , Aleix Sant$^{2*}$, Gerard I. Gállego$^2$, David Dale$^1$, Marta R. Costa-jussà$^1$} 

\address{Meta FAIR, Paris, France$^1$  \\ TALP Research Center, Universitat Politècnica de Catalunya, Barcelona, Spain$^2$ \\
         alastruey@meta.com, aleix.sant@estudiantat.upc.edu,\\
         gerard.i.gallego@upc.edu, daviddale@meta.com, costajussa@meta.com\\}

\abstract{
Speech-to-Speech and Speech-to-Text translation are currently dynamic areas of research. In our commitment to advance these fields, we present SpeechAlign, a framework designed to evaluate the underexplored field of source-target alignment in speech models. The SpeechAlign framework has two core components. First, to tackle the absence of suitable evaluation datasets, we introduce the Speech Gold Alignment dataset, built upon a English-German text translation gold alignment dataset. Secondly, we introduce two novel metrics, Speech Alignment Error Rate (SAER) and Time-weighted Speech Alignment Error Rate (TW-SAER), which enable the evaluation of alignment quality within speech models. While the former gives equal importance to each word, the latter assigns weights based on the length of the words in the speech signal. By publishing SpeechAlign we provide an accessible evaluation framework for model assessment, and we employ it to benchmark open-source Speech Translation models. In doing so, we contribute to the ongoing research progress within the fields of Speech-to-Speech and Speech-to-Text translation.
 \\ \newline  }

\begin{document}

\maketitleabstract

\section{Introduction}
\label{sec:intro}
Speech-to-text Translation (S2TT) and Speech-to-speech Translation (S2ST) refer to the task of converting spoken language into respectively written text or speech in a different language. These tasks are increasing their popularity, and can be used for applications such as subtitling videos in a different language, translating between languages that do not have a written form, and in general, ensuring seamless communication across people worldwide.

The initial approach to S2TT and S2ST involved the integration of distinct models, forming what is nowadays known as a cascade system \cite{758176}. This systems consist of an Automatic Speech Recognition (ASR) model that transcribes the spoken sentence, and a Machine Translation (MT) model that translates the sentence into another language. In the case of S2ST an additional speech synthesizer is needed, that is utilized to generate the corresponding speech from the translated text. However, recent advancements have led to the development of end-to-end models, that perform translation from speech to text or to speech, without requiring an intermediate transcription step, or a translated transcript. Known as direct Speech Translation systems, these models have quickly progressed, and currently, they can achieve state-of-the-art results comparable to those of cascade models \cite{Ansari2020, DBLP:journals/corr/abs-2106-01045}. Nevertheless, the performance of both cascade and end-to-end architectures remains far from optimal compared to text translation systems, indicating that research in these areas is still ongoing. 

\begin{figure}[h]
  \centering
  \includegraphics[width=6.5cm]{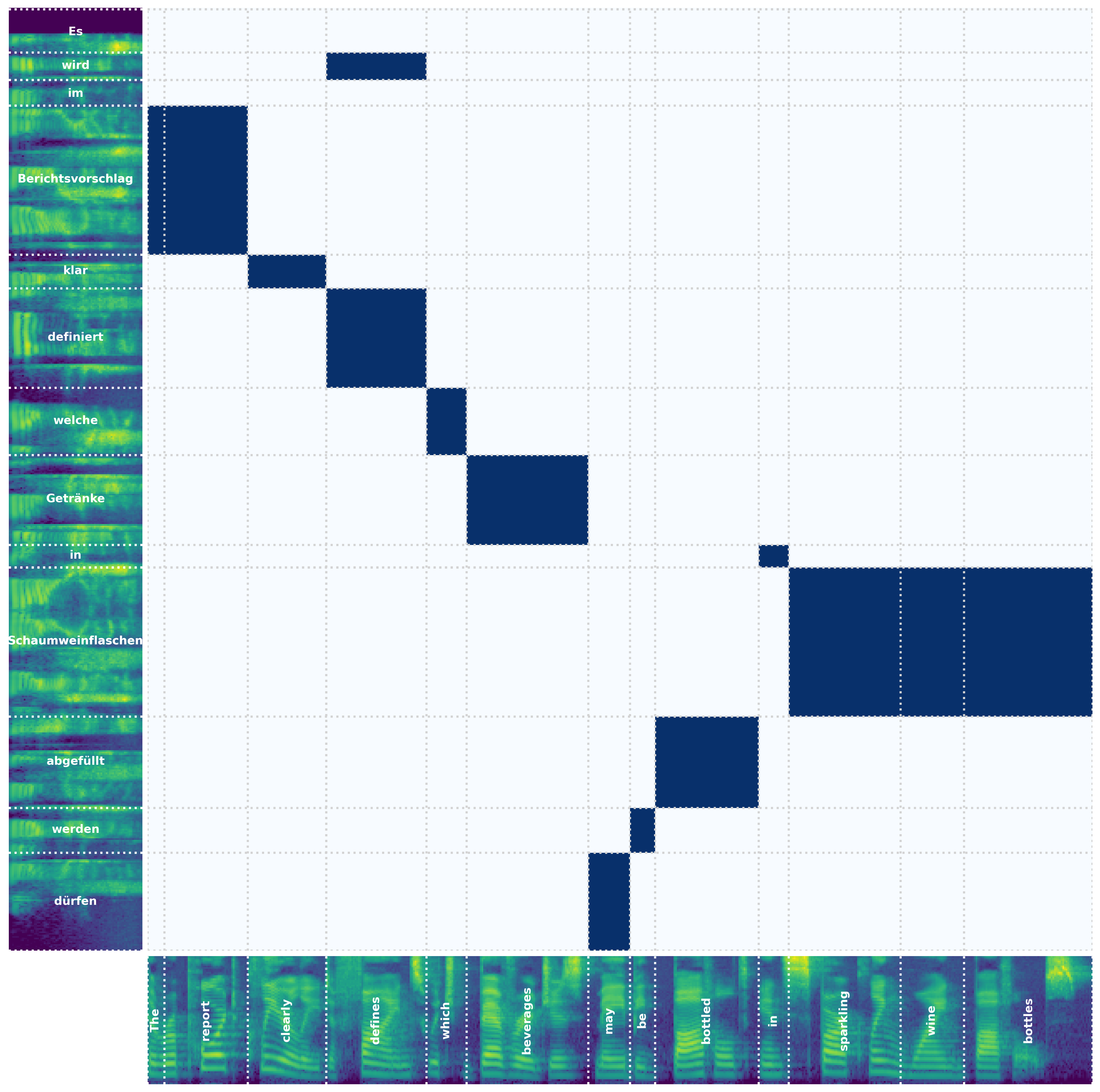}
  \caption{Example of a S2ST alignment in the Speech Gold Alignment dataset. }
  \label{fig:main_plot}
\end{figure}

The recent growth of end-to-end models and the shift in the field towards using them has raised the need to understand their inner workings. One related task is source-target alignment, which involves analysing how models use the provided source to make predictions, and whether they follow common human intuition in this process.

This alignment task has been widely explored in the context of text translation \cite{ghader-monz-2017-attention, ferrando-etal-2022-towards}. The task is commonly evaluated using Alignment Error Rate (AER) \cite{och-ney-2003-systematic}, a metric that measures the differences between a gold-standard alignment and a hypothesized one. For this aim, human-labeled alignment datasets have been published in the context of text translation, such as \cite{vilar06_iwslt} for translation between English and German.

In speech-related fields, little interpretability work regarding alignments has been done. Some previous studies have focused on analysing the self-attention in the encoder of speech recognition, \cite{zhang-etal-2021-usefulness,shim2022understanding} and speech translation \cite{alastruey-etal-2022-locality} systems. However, these models' decoder, and consequently, its alignment capabilities, have yet to be explored, potentially due to the absence of suitable datasets and metrics for evaluating the task in this setting.

In light of this, we introduce SpeechAlign framework\footnote{\url{https://github.com/mt-upc/speechalign}}, which serves as a solution to the stated lack of resources. SpeechAlign is formed of two core components: a novel dataset and an evaluation framework founded on our proposed metrics. 

The dataset, named Speech Gold Alignment, is specifically created to evaluate alignment in S2TT and S2ST. This dataset is an extension of the text translation gold alignment dataset introduced by \citet{vilar06_iwslt}. To create it, we employ a Text-to-Speech (TTS) model to generate synthetic speech for the sentences in the dataset.  The utilization of a TTS model offers a main advantage: apart from generating audio, it also provides timestamps denoting the beginning and end of each word. Annotating such timestamps would be very resource-intensive if using non-synthetic audios. Gathering the audios and the timestamps, we are able to build the Speech Gold Alignment dataset, formed of samples such as the one shown in Figure \ref{fig:main_plot}. 

In terms of metrics, we adapt the AER for the speech domain, introducing two novel metrics: Speech Alignment Error Rate (SAER) and Time-weighted SAER (TW-SAER). These metrics quantify the alignment error models have, with the key distinction that the former treats each word equally, and the latter factors in word durations.

To sum up, the main contribution of this paper is the release of SpeechAlign, a framework designed to simplify metrics computation using our dataset. Additionally, we employ this framework to benchmark various open-source models. Through these efforts, we aim to contribute to the exploration of alignments in the domain of speech translation.
\begin{figure*}[ht]
    \centering
    
    \begin{subfigure}{0.25\textwidth}
        \includegraphics[width=\linewidth]{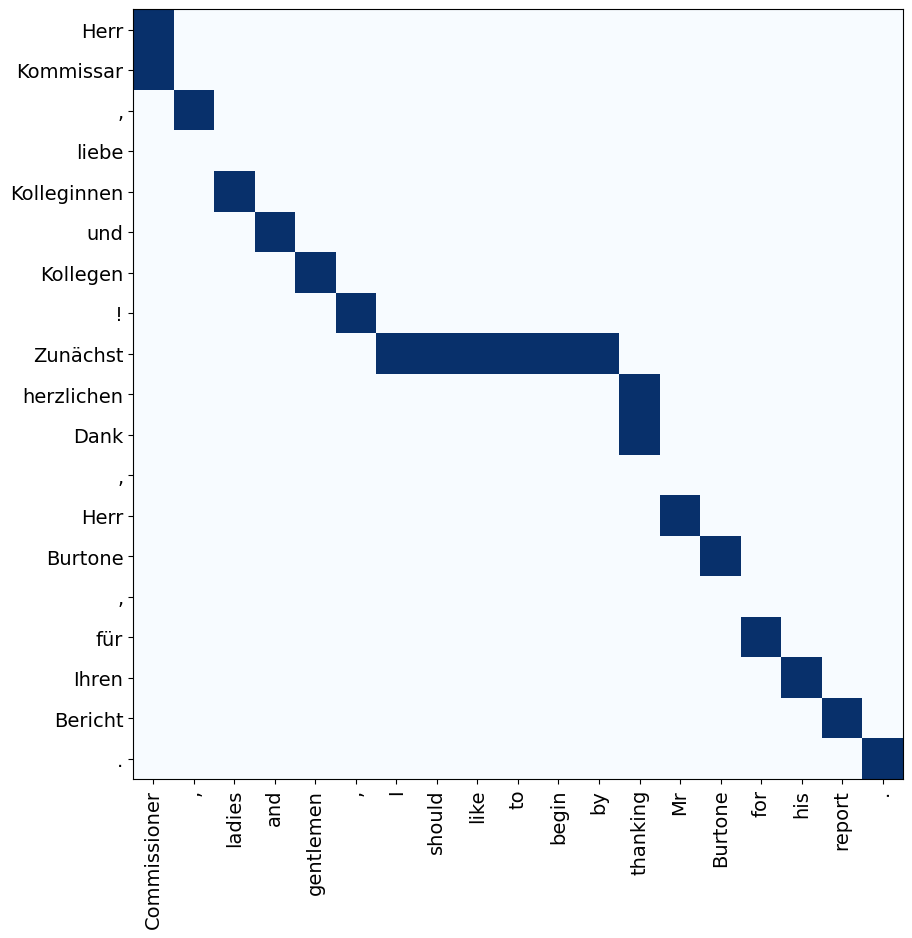}
        \caption{T2T Translation.}
        \label{fig:plot1}
    \end{subfigure}
    \hfill
    \begin{subfigure}{0.37\textwidth}
        \includegraphics[width=\linewidth]{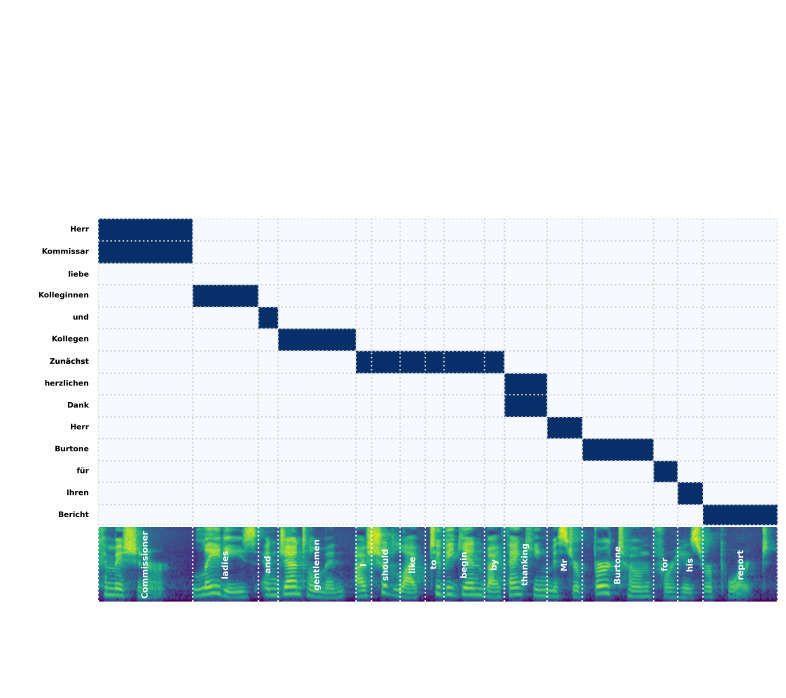}
        \caption{S2T Translation.}
        \label{fig:plot2}
    \end{subfigure}
    \hfill
    \begin{subfigure}{0.25\textwidth}
        \includegraphics[width=\linewidth]{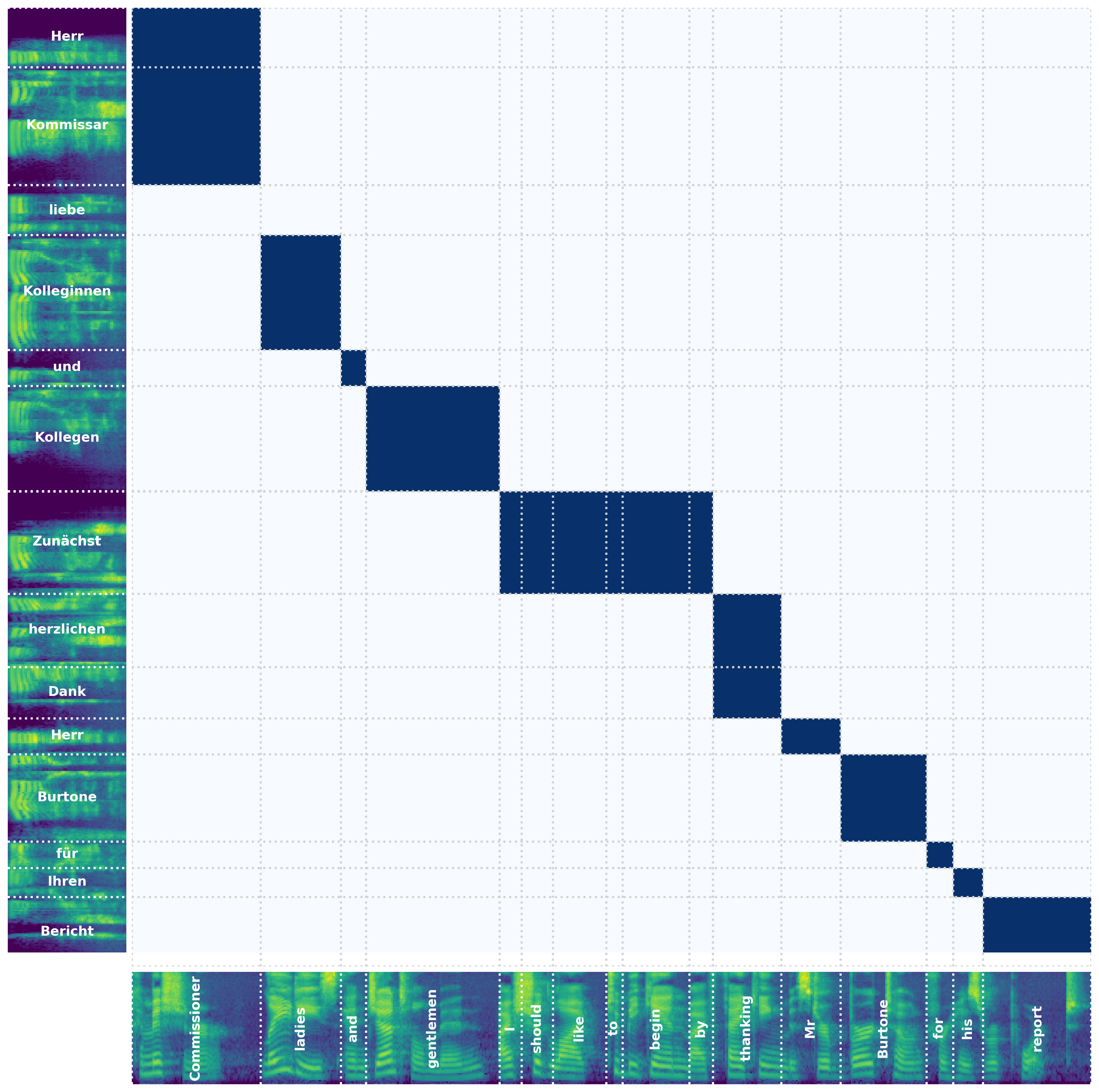}
        \caption{S2S Translation.}
        \label{fig:plot3}
    \end{subfigure}
    
    \caption{Original alignment by \cite{vilar06_iwslt} and our extensions.}
    \label{fig:all_plots}
\end{figure*}

\section{Related Work}
\label{sec:rel_work}
Over the past decades, considerable interest has been directed toward comprehending the alignment capabilities of text translation models. In this trajectory, both datasets and metrics have been developed to evaluate this task.

Numerous authors have published alignment datasets \cite{082f3feb-ebbf-3e12-92bb-e4f6356b7a40,vilar06_iwslt, kruijff-korbayova-etal-2006-annotation,  graca-etal-2008-building, macken-2010-annotation, holmqvist-ahrenberg-2011-gold} for the evaluation of alignments in translations in languages such as English, Spanish, German, Dutch, and Czech. In this work, we hone in on the dataset introduced by \citet{vilar06_iwslt}\footnote{\url{https://www-i6.informatik.rwth-aachen.de/goldAlignment/}} for text translation between English and German. This dataset comprises 508 paired sentences in the specified languages, along with precise information regarding the alignment of words between these two languages. These sentences are sourced directly from the EuroParl dataset \citelanguageresource{koehn-2005-europarl}, which contains transcripts and translations of speeches delivered in the European Parliament. We opt for this dataset due to its coverage of the English-German translation pair, which is extensively studied in the field of speech translation \cite{agrawal-etal-2023-findings}. Moreover, our work requires the generation of speech utterances for the sentences in the dataset. Focusing on well-resourced languages like English and German provides greater confidence in the quality of the speech generated by the TTS model.

As for metrics, a singular measure has predominantly been used to evaluate alignments. Alignment Error Rate (AER), introduced by \citet{och-ney-2003-systematic}, is a measure of alignment quality between a source sentence and its translation. It is calculated as the ratio of alignment errors, where an alignment error occurs when a unit in the translated sentence is not aligned with the correct unit in the source. The score is computed based on a manually annotated gold-standard alignment of a parallel corpus. Given a reference alignment, consisting of a set $S$ of “Sure”, unambiguous alignment points, and a set $P$ of “Possible”, ambiguous alignment points, with $S \subseteq P$, the AER of an alignment $A$ is defined to be:
 \begin{equation}
    \text{AER}(S, P; A) = 1 - \frac{|A \cap S| + |A \cap P|}{|A| + |S|}
\end{equation}

\section{Speech Gold Alignment Dataset}
\label{sec:dataset}
The dataset we introduce, \textit{Speech Gold Alignment}, extends the bilingual text alignment dataset presented by \citet{vilar06_iwslt} by adding speech utterances to each pair of English and German sentences. Additionally, for each audio file, the dataset contains a dictionary defining all the words in each sentence and their corresponding start and end time stamps, what gives us a text-to-audio mapping. Once we have this, we incorporate the gold alignment correspondences from the original dataset to obtain the alignments between speech segments. 

This augmented dataset, can either be considered as two distinct datasets supporting S2TT from English to German (and vice versa), or as a unified S2ST dataset by combining both S2TT alignments. In Figure \ref{fig:all_plots} we show the three different modalities of the dataset. Figure \ref{fig:plot1} shows a sample from the original dataset presented by \citet{vilar06_iwslt}, and Figures \ref{fig:plot2} and \ref{fig:plot3} show our extension for S2TT and S2ST settings respectively.

As part of the SpeechAlign framework, we publish a pipeline to prepare the dataset, following the steps that are described in section \ref{sec:construction}.

\subsection{Methodology}
\label{sec:construction}

The construction of this dataset can be divided into two primary steps. First, we employed the VITS model \cite{kim2021conditional} to generate synthetic speech for all the sentences, as detailed in section \ref{ssec: speech}. Subsequently, we aligned each word to its corresponding time interval in the produced speech signal, as explained in section \ref{ssec: align}. While integrating the datasets, we found specific cases where alignment was not immediate or direct. We address these complexities in section \ref{ssec: special_cases}.

\subsubsection{Speech Generation}
\label{ssec: speech}
To produce synthetic speech for the sentences in the Gold Alignment dataset, we employed the VITS model. This TTS system uses a phonemizer to obtain the phonemes corresponding to the input sequence. Then, to generate the speech output, the model uses a stochastic duration predictor that assigns a duration to each phoneme. The chosen duration is randomly sampled from each phoneme's durations distribution. By doing this, the model is able to synthesize natural speech and can generate different speech utterances for the same input text.

To build our dataset, we generated separate synthetic versions for the 508 sentences in both English and German. In English, we utilized LJ Speech \citelanguageresource{ljspeech17}, while for the German language, the Thorsten voice \citelanguageresource{muller_thorsten_2021_5525342} was employed. This task was done using the VITS model available through the Coqui toolkit \cite{coqui-tts}.

\subsubsection{Word-Audio Matching}
\label{ssec: align}
The Gold Alignment dataset constitutes a word-to-word alignment reference, to which we add our newly generated audios, product of VITS. Nevertheless, to achieve an alignment between speech intervals, we first need to establish a linkage between audio segments and words in the original dataset.

The approach followed to accomplish this starts by acquiring intermediate representations from VITS. Specifically, we gather the output generated by the phonemizer, which is the phonemized sentence, as well as the output of the duration predictor. This predictor creates a dictionary containing duration in integer units of each phoneme. With this information in hand, we perform a two-step matching procedure, that ultimately yields the mapping from audio to words, via the intermediate representation of phonemes:

\begin{enumerate}
    \item Phoneme-Word Matching. In this stage, we focus on aligning the phonemes with the words present in the original dataset.
    \item Phoneme-Audio Matching. In this phase, we establish a time mapping between the audio and its corresponding sequence of phonemes. 
\end{enumerate}

\begin{figure*}[h]
  \centering
  \includegraphics[width=0.9\textwidth]{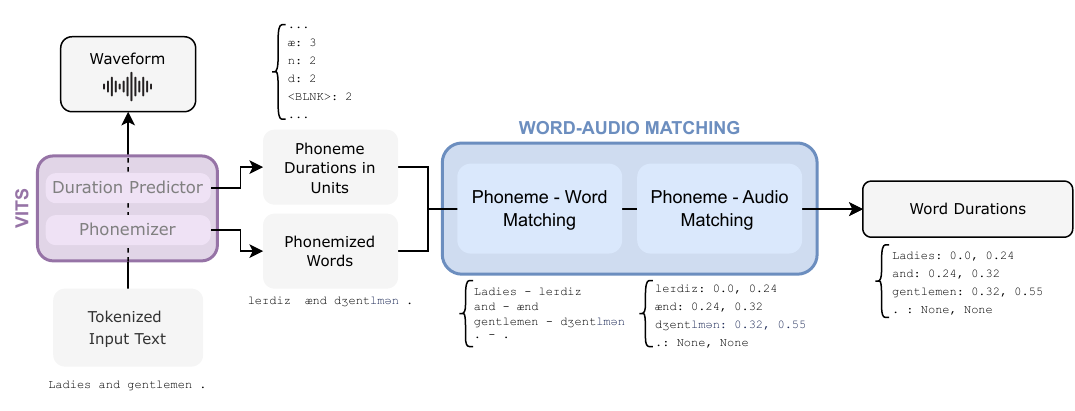}
  \caption{Pipeline used to generate the dataset.}
  \label{fig:pipeline}
\end{figure*}

Figure \ref{fig:pipeline} provides a visual representation of the sequential steps followed for deriving both the waveform and the alignment between words and audio, which constitute the dataset we present.

With the basic steps outlined, now we will dive deeper into the details of each of the phases to obtain the audio-word matching.

\subparagraph{Phoneme-Word Matching.}
The goal of this phase is to achieve a mapping between the sequence of phonemes extracted from the phonemizer and the sequence of words in the original dataset \cite{vilar06_iwslt}. To do so, we use blank spaces as delimiters for words in the phonemes sequence, and we monotonically map them with the sequence of words. It is important to note that the original dataset underwent tokenization through Moses, introducing some challenges in this process that are outlined in detail in Section \ref{ssec: special_cases}.

\subparagraph{Phoneme-Audio Matching.}
After obtaining the correspondence between words and phonemes, we now need to map phonemes to the audio. Ideally, the entire audio must be partitioned into separate time intervals, each containing the pronunciation of a single word. To accomplish this, it is necessary to compute the overall duration of each individual word.

To compute the total duration of each word, we take the output of the duration predictor and sum the duration in units of all the phonemes belonging to a same word. As previously stated, blank spaces are used as delimiters between words in the phoneme transcription. Consequently, the duration assigned to a blank space is equally distributed and added to the neighboring words, both preceding and succeeding the blank space. The same approach applies to units attributed to punctuation marks, that we decided not to include in our alignment dataset given that they cannot be found explicitly in speech utterances. 

Next, our objective is to establish the corresponding word duration in seconds based on their duration in units. To achieve this, we divide the total length of the audio by the aggregate duration in units of all the phonemes in the sentence. This computation establishes a correlation between VITS duration units and the equivalent time in seconds. Using this derived relationship, we convert the word durations from units to seconds and find the start and end times for each word.

\subsubsection{Special Cases}
\label{ssec: special_cases}
After the two phases of the dataset construction we perform a manual revision of the generated data and encounter some special challenges that need special handling. 

\subparagraph{Phonemic Fusion.} In the majority of instances, phonemized words align with the original text words, primarily through sentence segmentation using blank spaces. Nevertheless, in certain cases the phonemizer merges adjacent words during phonetic transcription, creating what we name as \textit{phonemic fusion}. This occurrence is primarily observed in short English words such as prepositions, articles, and pronouns, which are pronounced seamlessly without pauses. Table \ref{tab:phonemizer_issues} provides examples of this phenomenon. In such particular instances, we first determine the combined duration of these merged words and subsequently distribute the total time proportional to the length among the constituent words. While this approach may not be entirely precise, we believe the approximation is enough, given its applicability to very short words and few cases.

\begin{table}[ht]
    \centering
    \small
    \renewcommand{\arraystretch}{1.2}
   \begin{tabular}{l}
    \toprule
     \textbf{Phonemic Fusion} \\
    \hline
     Words: I am                    \\ 
     Phonemes: \textipa{/aIam/}\\ 
     \hline
     Words: of the\\ 
     Phonemes: \textipa{/@vði/}\\ 
     \hline
     Words: as it is\\ 
     Phonemes: \textipa{/æzItIz/}\\ 
     \hline
     Words: that the\\ 
     Phonemes: \textipa{/ðætði/} \hphantom{ssssssssssssssssssssssss}\\ 
    \bottomrule
    \end{tabular}
    \centering
    \renewcommand{\arraystretch}{1.2}
   \begin{tabular}{l}
    \toprule
    \textbf{Phonemic Fragmentation} \\
     \hline
     Words: 124 \\ 
     Phonemes: \textipa{/wVn hVndr@d twEntI fO/} \hphantom{sssssss} \\ 
     \hline
     Words: 34\%\\ 
     Phonemes: \textipa{/ðEtI fO p@sent/}\\ 
     \hline
     Words: 1996 \\ 
     Phonemes: \textipa{/naIntIn naIntI sIks/}\\ 
    \bottomrule
    \end{tabular}
  \caption{Examples of the special cases encountered when aligning words and their phonemization.}
  \label{tab:phonemizer_issues}
\end{table}

\subparagraph{Phonemic Fragmentation.}
Furthermore, we have encountered a contrasting phenomenon in comparison to \textit{phonemic fusion}. The phonemizer carries out a normalization process on the text before phonemization. Occasionally, this normalization procedure results in the conversion of single words into multiple words – a phenomenon we refer to as \textit{Phonemic Fragmentation}. This behavior is particularly noticeable in cases involving numbers, percentages, years, and similar elements. To address this matter, we aggregate the durations of all the split words and attribute the total duration to the original solitary word. 

\paragraph{Possesives ('s)}
The original Gold Alignment dataset does not provide alignments between natural sentences, but for sentences tokenized with Moses. However, VITS works on natural text, and this missmatch creates some difficulties along the matching process. This is the case of words such as "Parliament's", that is considered a single word when dealing with VITS ("Parliament's"  $\rightarrow$ \textipa{/pal@m@nts/}) , but it is actually two different words with independent alignments in the original dataset ("Parliament 's"), due to Moses tokenization.

This is a case of \textit{Phonemic Fusion}  (and it's addressed as such). However, unlike previously shown cases caused by the phonemizer, this fusion stems from the tokenization in the original dataset.

\subparagraph{Percent Sign (\%)}
A similar behaviour arises when dealing with percent signs. These signs appear alongside numbers in natural text ("34\%"), but in the Gold Alignment dataset, they're separate tokens due to Moses tokenization ("34 \%"). However, as illustrated in Table \ref{tab:phonemizer_issues}, percents are a case of \textit{Phonemic Fragmentation}, with the phonemizer breaking this construction into multiple phonemized words ("34\%" $\rightarrow$ \textipa{/ðEtI fO p@sent/}).

In this particular cases of Phonemic Fragmentation, we aim to separate the expanded phonetic text into two segments: the first containing phonemized words associated with the number (\textipa{/ðEtI fO/}), and the last containing the phonemized word corresponding to the percent (\textipa{/p@sent/}). In this instances, the merging of time intervals encompasses all words except the final one in the expansion.

\subparagraph{German Phonemizer}
In our utilization of the German phonemizer, we have noticed that it occasionally produces inaccurate phonetic transcriptions for certain single input words.  These inaccuracies tend to occur with special symbols (e.g., \textit{"\%", "/"}), years (e.g., "1996"), acronyms (e.g., "EU", "Nr"), compound nouns (e.g., "EU-Staate"), among others.  To rectify these inaccuracies in phonetic transcription, we have replaced specific words in the input sentences with their expanded and "spoken" format ("EU" $\rightarrow$ "E U", "1996" $\rightarrow$ "nineteen ninety six"). This adjustment assists the phonemizer in producing more accurate transcriptions.

\subsection{Dataset Quality Assessment}
\label{ssec:assesment}
Within this section, we aim to examine the quality of the synthetic audio produced by VITS. We conduct an assessment comparing EuroParl ST \citelanguageresource{9054626} test set and our own synthesized data, which is also derived from a subset of the EuroParl dataset \citelanguageresource{koehn-2005-europarl}. With this aim, we evaluate the performance of the Whisper Tiny model \cite{radford2022robust} on the task of speech recognition on these two datasets.  This strategy allows us to understand the implications of using synthetic audio without the influence of content domain. We choose to perform this evaluation in the setting of speech recognition, and not in translation, because of the simplicity of the former due to its monotonic alignment process. This ensures that the overall model performance and the complexity of the task are less likely to influence the results. We have opted to conduct this evaluation using the smallest Whisper model. Our hypothesis behind this choice is that if no issues arise in the smallest model, they are unlikely to manifest in larger models.

In Table \ref{tab:domain}, we present the Word Error Rate (WER) results obtained on both datasets, and we observe that our synthetic audios result in a lower WER than standard EuroParl ST dataset. Furthermore, we notice a disparity in performance between the German and English synthetic data. This discrepancy may stem from differences in the underlying VITS models, that are trained on distinct datasets for each language. This quality discrepancy between the German and English outputs was confirmed during the manual inspection of the generated speech. Despite this variance, it's important to point out that WER for German remains below the threshold established by Europarl ST, which serves as a quality reference in our study.  Consequently, we can conclude that the synthesized data does not pose a problem and appears to be easily handled by the models in both languages, possibly due to the clarity of the generated audios compared to European Parliament recordings.

\begin{table}
\small
\centering
\begin{tabular}{lcc}
\toprule
\textbf{Dataset} & \textbf{Language} &  \textbf{WER}  \\
\toprule
EuroParl ST & En & 29.7 \\
Speech Gold Alignment & En & 3.9\\
EuroParl ST & De & 31.0 \\
Speech Gold Alignment & De & 23.1 \\
\bottomrule
\end{tabular}
\caption{Quality assessment results.}
\label{tab:domain}
\end{table}

\section{Proposed Evaluation}
\label{sec:metrics}
The objective of this section is to define an evaluation procedure and metrics that are able assess models' ability to establish source-target alignments. To analyse this capability, our focus is on the contribution maps generated by the models. These maps indicate the relationship between source and target tokens, such that the contribution of a source token to a target one is always a non-negative value, and that the sum of contributions from all source tokens to a target token must equal 1 (i.e. attention weights or more advanced interpretability methods \cite{kobayashi-etal-2021-incorporating,ferrando-etal-2022-towards}). Then, to measure the alignments, we build new metrics around the intuition of the Alignment Error Rate (AER) score, initially introduced by \cite{och-ney-2003-systematic} and defined in section \ref{sec:rel_work}. However, extracting the alignments from the contribution map and adapting AER for speech sequences is not straightforward process.

\label{appx:prepro}
\begin{figure*}[h]
  \centering
  \includegraphics[width=0.99\textwidth]{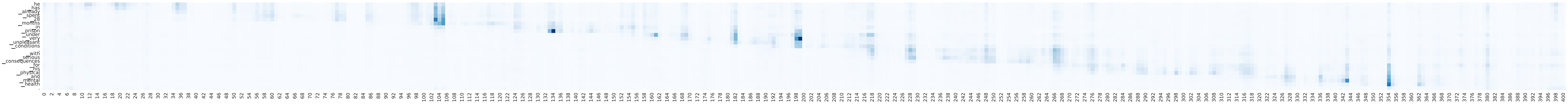}
  \caption{Example of the token-to-token attention weights of a S2TT decoder layer on Whisper Small.}
  \label{fig:pre}
\end{figure*}

\begin{figure}[h]
  \centering
  \includegraphics[width=0.48\textwidth]{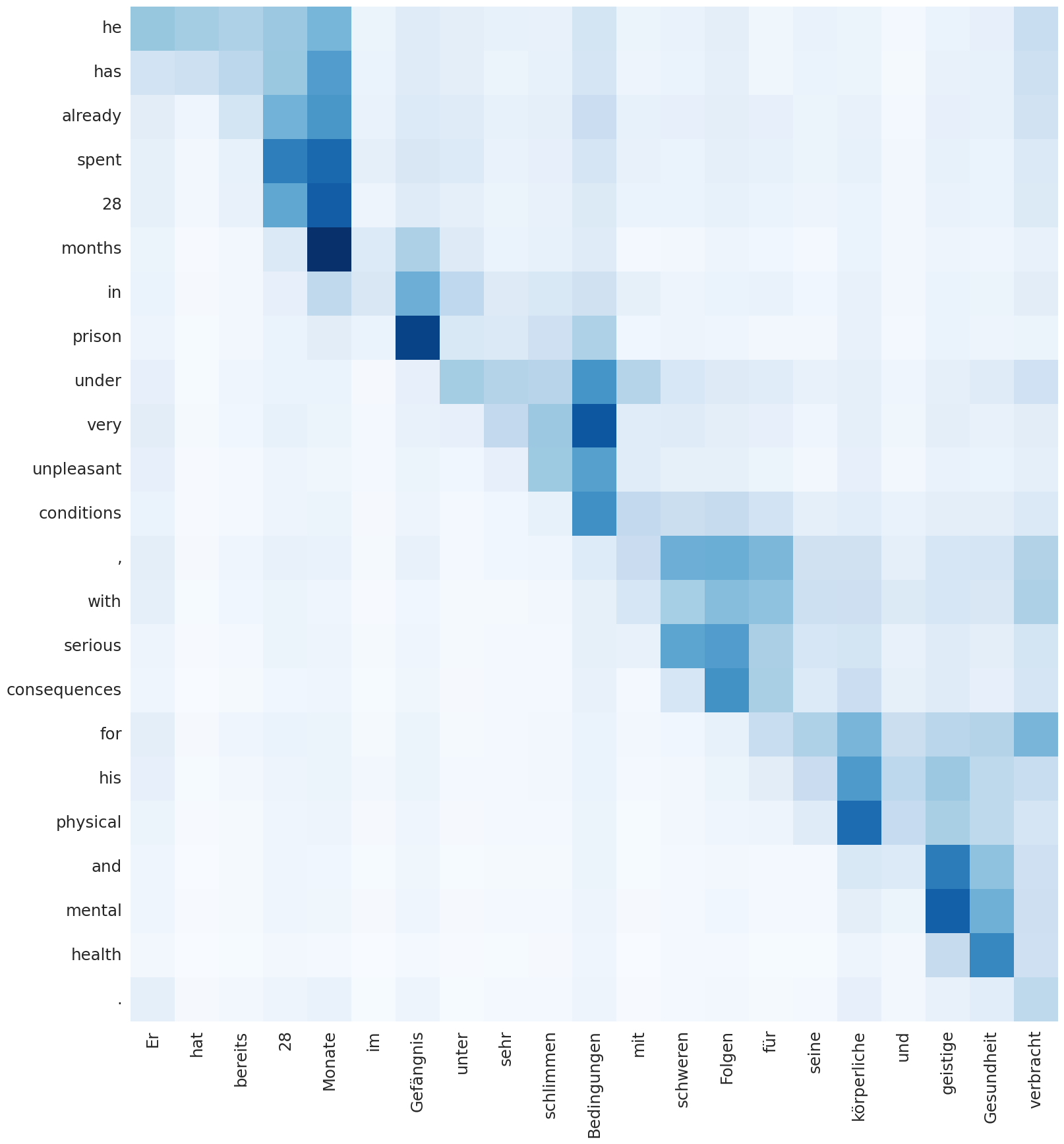}
  \caption{Example in Figure \ref{fig:pre} after the preprocessing, obtaining a word-to-word contributions map.}
  \label{fig:post}
\end{figure}

\begin{algorithm}[!h]
\DontPrintSemicolon
\footnotesize
\caption{\small Contributions Preprocessing}\label{alg:cap}
\KwIn{\;$C\_t2t$: token-to-token contribution matrix, \\ $src$: source words \& durations, \\ $tgt$: tgt words \& durations}
\KwOut{\\$C\_w2w$: word-to-word contribution matrix}\;
\SetInd{0.65em}{0.1em}
$max\_duration\_src \gets src[-1][end]$\;
$max\_tokens\_src \gets C\_t2t.shape[1]$\;
\For {$word, word\_idx \gets src$}{
    $s\_time \gets src[word][start]$\;
    $e\_time \gets src[word][end]$\;
    $s\_token \gets ceil(s\_time*max\_tokens\_src/max\_duration\_src) $\;
    $e\_token \gets floor(e\_time*max\_tokens\_src/max\_duration\_src)$\;
    $C\_w2t[:, word\_idx] \gets \text{sum}(C\_t2t[:, s\_token:e\_token], dim=1)$\;
}
$max\_duration\_tgt \gets tgt[-1][end]$\;
$max\_tokens\_tgt \gets C\_t2t.shape[0]$\;
\For {$word, word\_idx \gets tgt$}{
    $s\_time \gets tgt[word][start]$\;
    $e\_time \gets tgt[word][end]$\;
    $s\_token \gets ceil(s\_time*max\_tokens\_tgt/max\_duration\_tgt) $\;
    $e\_token \gets floor(e\_time*max\_tokens\_tgt/max\_duration\_tgt)$\;
    $C\_w2w[word\_idx, :] \gets \text{avg}(C\_w2t[s\_token:e\_token, :], dim=0)$\;
}

\end{algorithm}

\subsection{Preprocessing}

The metric of AER assesses the error rate between a hypothesis and a target alignment. Hence, to compute this score, we need a gold alignment dataset. In most text alignment datasets, such as the one we extend \cite{vilar06_iwslt}, these alignments are provided as word-to-word relations. Consequently, the hypothesis alignment needs to be structured in a word-to-word format too. However, in speech settings, the system input tokens correspond to frames of a spectrogram or ranges of a waveform. As a consequence, the contribution maps usually extract token-to-token interactions, being each token a speech frame. Thus, a conversion process is necessary to derive word-to-word alignments from a tokens-to-tokens contribution map, and consequently being able to evaluate the alignment to obtain an AER score.

Nonetheless, a similar challenge is faced in the setting of text translation, where tokens are often sub-words rather than complete words. In this case, the conversion from tokens to words involves a two-step process.When dealing with sub-words in the source, their contributions are aggregated by summing them together. This approach is rooted in the principle that the combined contribution of two tokens to a target is the sum of their individual contributions. Handling sub-words in the target sequence proves to be more complex. Each token has a distinct distribution of contributions across the source. To address this, the average of each sub-word distribution is computed. By following this approach, we are able to effectively establish the alignment between words despite the presence of sub-word units. 

In the case of speech, we propose to employ a similar approach when aggregating tokens from each word, in order to obtain a word-to-word contributions plot. Leveraging our dataset, which provides details into the correspondence between segments of input/output audio and individual words, we define which tokens correspond to each word under with the assumption of a linear relation between tokens and audio, and dismissing any overlap.  Doing this allows us to employ a similar approach to the one used for merging sub-words, but this time, we apply it to the set of all tokens linked to a single word. Given a contributions map $C$ where $c_{i,j}$ is the contribution of the  $j$-th source token to the $i$-th prediction, the resulting word-to-word contributions map is computed using our dataset as shown in Algorithm \ref{alg:cap}. In Figures \ref{fig:pre} and \ref{fig:post} we show an example of a contributions map before and after the preprocessing.

Following this conversion and before computing the alignment scores, we derive the hard alignments. This is accomplished by aligning each target word with the source word that has the highest contribution.

\subsection{Speech Alignment Error Rate}

Once we have the hard alignments, we define the Speech Alignment Error Rate (SAER) in the same manner the AER is defined. This is, given a set $S$ of unambiguous alignments, a set $P$ of ambiguous alignment and a set $A$ of hypothesis alignment:

\begin{equation}
    SAER = 1 - \frac{|A \cap S| + |A \cap P|}{|A| + |S|}
\end{equation}

Note that while the equation remains identical to that in the original AER case, the definitions of $A$, $S$, and $P$ change due to the preprocessing required to derive them.

Furthermore, it's important to note that SAER doesn't fully address a key aspect in the speech setting – the noticeable disparity in the number of different tokens that form each word, which corresponds to audio durations. Instead, when computing SAER each word contributes equally to the final score, regardless of its duration. Hence, tokens that correspond to short words are assigned a higher weight in the metric than those that correspond to longer words. This differs from the model perspective, where each token holds equal importance.

\subsection{Time-Weighted SAER}
\begin{figure}[h]
  \centering
  \includegraphics[width=0.48\textwidth]{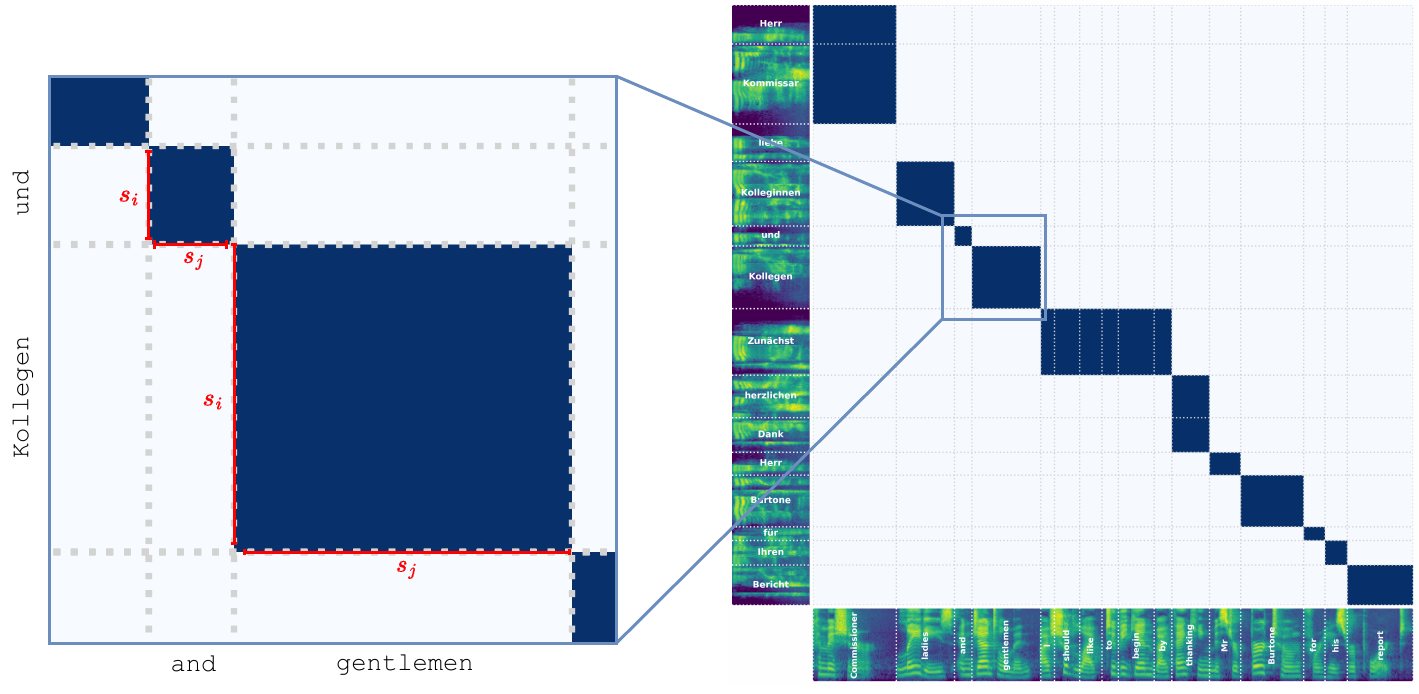}
  \caption{TW-SAER weights.}
  \label{fig:tsaer}
\end{figure}

To address the limitations of the SAER, we define the Time-weighted SAER, a metric that accounts for the variability in word durations. To do so, we introduce a new element – the incorporation of a weight for each alignment. These weights are defined using the area of each alignment, as shown in Figure \ref{fig:tsaer}, and defined as follows:

\begin{equation}
    w_{i,j} =
    \begin{cases}
        s_j \cdot s_i & \text{if S2ST}\\
        s_j \cdot 1 & \text{if S2TT}
    \end{cases}
\end{equation}

where $w_{i,j}$ is the weight of an alignment between the $j$-th source word and the $i$-th target word, and $s_i$, $s_j$ is the duration in seconds of these words respectively. Therefore, given a set set $S$ of unambiguous alignments and a set $P$ of ambiguous alignment, the TW-SAER is defined as the sum of areas of the alignments in $A \cap S$ plus the sum of areas in $A \cap P$, divided by the total alignment area of $A$ and $S$:

\begin{equation}
\small
    TW-SAER = 1 - \frac{\sum_{i,j \in A \cap S}w_{i,j} + \sum_{i,j \in A \cap P}w_{i,j} }{\sum_{i,j \in A}w_{i,j}  + \sum_{i,j \in S}w_{i,j}}
\end{equation}

By including the weights we account for the temporal duration of each word within the audio, refining our evaluation process. Note that SAER and TW-SAER are equivalent if $w_{i,j} = 1\hspace{2pt}\forall i, j$.

\begin{table*}
\centering
\begin{tabular}{lc|ccc}
\toprule
 \textbf{Size} & \textbf{Parameters
}  &  \textbf{SAER}(\%, $\downarrow$) &  \textbf{TW-SAER}(\%,$\downarrow$) & \textbf{BLEU}($\uparrow$) \\
\toprule
Tiny  & 39M  &  75.3 & 70.1 &  3.6  \\
Base & 74M &   72.9 & 67.8 & 8.4 \\
Small & 244M  &  70.7 & 65.7 &  15.4 \\
Medium & 769M &  69.5 & 64.1 & 20.2 \\
Large & 1.55B &  68.9 & 63.5 & 22.1 \\
\bottomrule
\end{tabular}
\caption{Benchmarking of different sizes of Whisper models on De-En S2TT.}
\label{tab:results}
\end{table*}

\section{SpeechAlign}
The main contribution of this paper is the release of SpeechAlign, an accessible open-source framework that encompasses the Speech Gold Alignment dataset presented in section \ref{sec:dataset} and the SAER and TW-SAER metrics defined in section \ref{sec:metrics}. This tool seamlessly handles raw token-to-token alignment maps and computes both proposed alignment error rates. This framework is versatile, and can be used in attention weights or more sophisticated contribution maps. The pipeline starts by taking the given contribution maps and converts them into word-to-word equivalents. To achieve this, the alignment dataset is utilized to account for varying word durations. Following this conversion, we derive the hard alignments. The outcome is a definitive set of hypothesis alignments, that are used to compute both SAER and TW-SAER scores.

To enhance the comprehension of the process, we include a notebook for visualization of the alignments and contributions maps. This tool can be used to visualize token-to-token and the extracted word-to-word representations, and also the obtained hard alignments. By publishing this framework, we aim  to facilitate the use of our dataset by other researchers.

Finally, using SpeechAlign, we benchmark some S2TT models. For simplicity, we decide to analyze alignments based on models' cross-attention weights. We decide not benchmark the S2ST task due to the current lack of open-source models, being the recently published SeamlessM4T \cite{seamlessm4t2023} the only one available as of now. This model comprises two consecutive Transformers, each containing its own decoder. Consequently, it presents significant challenges in terms of obtaining a contributions map based on attention weights, and developing further interpretability methods lies beyond scope of this paper.

\paragraph{Models Benchmarking}

Table \ref{tab:results} presents an evaluation of various sizes of the Whisper model \cite{radford2022robust} on De-En S2TT. Each model's performance is assessed through the BLEU score on our test set, and the SAER and TW-SAER. The latter are computed on the attention weights of each decoder layer, and in Table \ref{tab:results} we report the best obtained score. This analysis uncovers a correlation between the performance metrics and the alignment score. This correlation is also observed to align with the model's size. 

\section{Conclusion}
In conclusion, this paper introduces SpeechAlign, a framework to evaluate alignment in speech models. SpeechAlign has two main components. Firstly, we've created the Speech Gold Alignment dataset, being the first of its kind and created to address the lack of suitable evaluation data for the task. Secondly, we have presented the two first evaluation metrics for speech alignment, Speech Alignment Error Rate (SAER) and Time-weighted Speech Alignment Error Rate (TW-SAER), to assess how well speech models perform on the alignment task. SpeechAlign provides an accessible way to evaluate speech models, and we have used it to benchmark various open-source models.

\nocite{*}
\section{Bibliographical References}\label{sec:reference}

\bibliographystyle{lrec-coling2024-natbib}
\bibliography{lrec-coling2024-example}

\end{document}